\pdfoutput=1
\documentclass[11pt]{article}

\usepackage[final]{acl}

\usepackage{times}
\usepackage{latexsym}
\usepackage{hyperref}
\usepackage{xurl}
\usepackage{caption}
\usepackage{subcaption}
\usepackage{natbib}

\usepackage{svg}
\usepackage{float}

\usepackage[T1]{fontenc}
\usepackage[utf8]{inputenc}

\usepackage{microtype}

\usepackage{inconsolata}

\usepackage{graphicx}

\title{Instructions for *ACL Proceedings}

\usepackage{times}
\usepackage{latexsym}
\usepackage{enumitem}
\usepackage{natbib}
\usepackage[T1]{fontenc}
\usepackage[utf8]{inputenc}

\usepackage{microtype}

\usepackage{inconsolata}

\usepackage{graphicx}

\usepackage{times}
\usepackage{latexsym}

\usepackage{booktabs}
\usepackage{siunitx}
\usepackage{amsmath,amssymb,amsthm}
\usepackage{multirow}
\usepackage{amsmath}
\usepackage{cleveref}
\usepackage{tikz}
\usepackage{array}
\newcolumntype{C}[1]{>{\centering\arraybackslash}p{#1}}
\usepackage{microtype}
\usepackage{tabularx}

\title{Subword Segmental BabyLMs:\\Learning to Tokenise for Data-Efficient Language Modelling}
\title{Subword Segmental BabyLMs:\\Learning to Tokenise for Sample-Efficient Pretraining}
\author{Francois Meyer\\
  Department of Computer Science\\
  University of Cape Town\\
  \texttt{francois.meyer@uct.ac.za}}

\begin{document}
\maketitle

\begin{abstract}

In the standard LM training pipeline, subword tokenisation is applied as a preprocessing step. Subword segmental language modelling is an alternative paradigm in which tokenisation is learned during training, allowing the model to discover subword units that optimise its training objective. In this paper, we present our submission to the 2026 BabyLM Challenge, for which we develop two new subword segmental LMs: SubSegGPT and SubSegDeBERTa. SubSegGPT is a decoder-only model that learns tokenisation during autoregressive pretraining.  SubSegDeBERTa is an encoder-based model that jointly learns to generate and tokenise masked words. We train both for the \textsc{Strict} and \textsc{Strict-small} tracks. Our top submission to \textsc{Strict} is SubSegDeBERTa, which achieves notable gains in zero-shot evaluation. Our top submission to \textsc{Strict-small} is SubSegGPT, which outperforms tokenisation-based baselines. Our results show that learnable subword tokenisation can improve sample-efficiency for BabyLM pretraining. We analyse the subword learning dynamics of our models and find that tokenisation gradually converges on subword units that balance morphological alignment and fine-grained segmentation.

\end{abstract}

\section{Introduction}
\label{sec:introduction}

% \begin{figure}[t]
%  \vspace{-0.2cm}
%   \includegraphics[trim={0.25cm 0.25cm 0 0cm},clip,width=\linewidth]{figures/intro_scatter.pdf}
%   \vspace{-0.6cm}
%   \caption{Geometric properties of final-layer token representations in Llama 3.1 8B (see Figures~\ref{fig:full_final_layer_cosine} and~\ref{fig:full_final_layer_isoscore} in the appendix for all models). Lower-resource languages exhibit greater representational degeneration. } 
%   \vspace{-0.4cm}
%   \label{fig:llama8b_final_layer_cosine}
% \end{figure}

% Whhat do I want to convey here?
% Fixed tokenisation is the standard, but not great. Also not developmentally plausible.
% Another option is subword segmentral modelling.
% SubSegGPT and SubSegDeBERTa - what they are.
% What we do.
% Results and analysis
Current language model (LM) training is developmentally implausible in many ways, one of which is its reliance on subword tokenisation. 
%One of the many ways in modern language model (LM) training is developmentally implausible, is its reliance on subword tokenisation. 
The subwords produced by standard tokenisers 
%are linguistically implausible units in that they
do not reliably align with morpheme boundaries %\citep{bostrom-durrett-2020-byte}.% , 
\citep{batsuren2024evaluatingsubwordtokenizationalien}. 
Moreover, the dominant paradigm of applying a fixed tokeniser during preprocessing does not mirror child language learning, as humans are not born pre-equipped with a fixed lexical vocabulary. Instead, children incrementally learn to segment speech into meaningful units during language acquisition \citep{speech_segmentation_review}. 

\begin{figure*}[t]
        \centering
       \includegraphics[clip, trim=0 0.3cm 0 0.5cm, width=\linewidth]{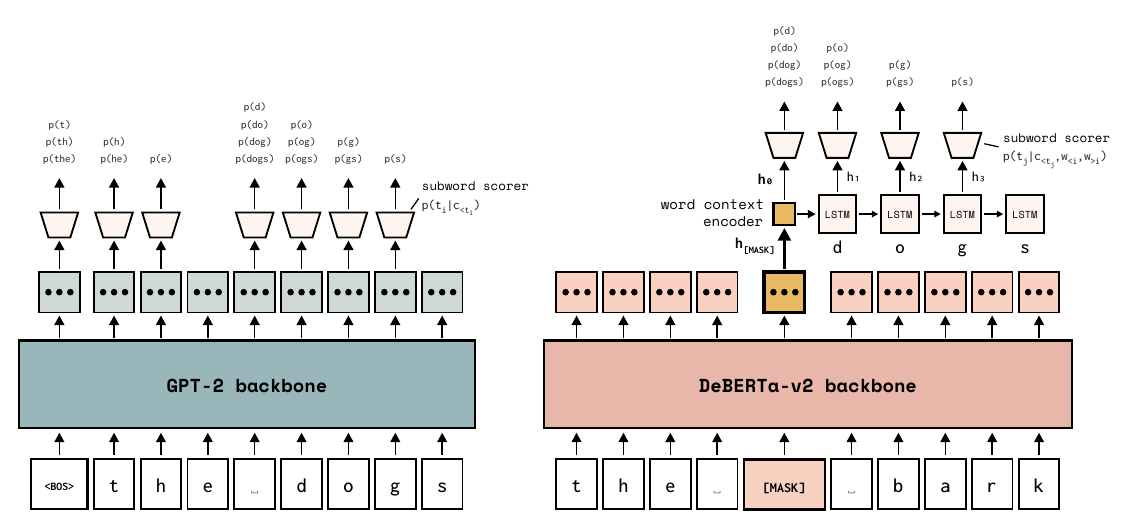}
    \vspace{-0.35cm}
    \caption{The architectures of SubSegGPT (left) and SubSegDeBERTa (right). Both encode character-level context, which is passed to a subword segment scorer to condition next-subword probabilities. Subword probabilities are passed to a dynamic programming algorithm that marginalises over all possible tokenisations.  } 
    \label{fig:combined}
    \vspace{-0.25cm}
\end{figure*}

From a more practical, performance-driven perspective, standard NLP tokenisation poses problems for data-efficient modelling. 
The algorithms behind tokenisers like BPE \citep{sennrich-etal-2016-neural} and ULM \citep{kudo-2018-subword} learn subword boundaries based on frequency-based objectives, with no guarantee that the resulting units are optimal for LM learnability.  
Given sufficient training data, neural LMs are robust to sub-optimal tokenisation. % can overcome
However, in small data settings, this may further compound the difficulty of learning generalisable linguistic representations. 

By restricting tokenisation to preprocessing, LMs are bound to a pre-determined tokenisation scheme.  %throughout training. 
Alternatively, tokenisation can be cast as a learnable component of language modelling, to be continually optimised \emph{during} training. This is the motivation behind subword segmental modelling \citep{meyer-buys-2022-subword}, which unifies tokenisation and language modelling in an end-to-end trainable framework. Instead of determining subword boundaries before training, subword segmental modelling marginalises over all possible tokenisations of a sentence and allows the model to learn which subword units optimise its LM training objective. % (see Section~\ref{subsec:background_subseg} for technical details).

In this paper, we present our submission\footnote{Models: \href{https://huggingface.co/collections/francois-meyer/babylm-2026-models}{\tt francois-meyer/babylm-2026-models}} to the 2026 BabyLM Challenge, for which we develop\footnote{Code: \href{https://github.com/francois-meyer/subseg-babylms}{\tt github.com/francois-meyer/subseg-babylms}} two new \textbf{Sub}word \textbf{Seg}mental LMs: SubSegGPT and SubSegDeBERTa.
SubSegGPT is a decoder-only subword segmental architecture with a GPT-2 \citep{gpt2} backbone. It combines subword segmental modelling with more modern, GPT-style conventions in decoder LM design and implementation. %architectures. 
SubSegDeBERTa is a novel, encoder-based architecture and the first instantiation of subword segmental modelling for masked language modelling. It jointly learns to generate and tokenise masked words during training, conditioned on bi-directional context from a DeBERTa-v2 backbone \citep{he2021deberta}.

We train our models for the \textsc{Strict} (100M words) and \textsc{Strict-small} (10M words) tracks and evaluate on the official evaluation pipeline, which includes zero-shot evaluations, finetuning tasks, and human likeness tests. We compare our models to equivalent tokenisation-based BabyLM baselines, based on respectively GPT-2 and DeBERTa-v2, to isolate the effect of learning tokenisation over fixed tokenisation.
% Both of these models learn subword boundaries during training, allowing them to opptimise subword tokenisation for developmentallyt plausible training. We argue that this is more aligned with how humans learn language -- it's more developmentally plausible -- than fixed subword tokenisation. 

% On both tracks, SubSegGPT and SubSegDeBERTa outperform tokenisation-based baselines on zero-shot and finetuning evaluations. 
In the \textsc{Strict} track, SubSegDeBERTa and SubSegGPT outperform tokenisation-based baselines on average across NLP tasks. SubSegDeBERTa is our strongest submission to \textsc{Strict}, improving average zero-shot performance by 3.16 points over GPT-2. % and ranking among the top models on the official BabyLM leaderboard. 
In \textsc{Strict-small}, SubSegGPT outperforms  baselines, but SubSegDeBERTa does not. 
Our two models are complementary: SubSegDeBERTa is highly competitive in the \textsc{Strict} track, while SubSegGPT is better suited to the more data-constrained \textsc{Strict-small} setting.
On human likeness tasks, our models exhibit little resemblance to human language acquisition and are outperformed by tokenisation-based baselines. 
% consistently outperforms SubSegDeBERTa and achieves performance competitive with the reigning BabyLM Challenge champion, GPT-BERT \citep{charpentier-samuel-2024-bert}, on a few evaluation tasks. 

Lastly, we analyse the subword learning dynamics of SubSegGPT and SubSegDeBERTa. We compare learned subword units over the course of training, tracking fertility and morpheme boundary F1. Both models undergo an initial period of rapid tokenisation change, followed by stabilised learning trajectories that converge on higher fertility (shorter subwords) and limited alignment with morpheme boundaries, which we support with a qualitative analysis of tokenised child-directed speech.

% This must be in the first 5 lines to tell arXiv to use pdfLaTeX, which is strongly recommended.

\section{Background}
\label{sec:related_work}

%\subsection{Speech Segmentation in Child Language Acquisition}

% Children dynamically adjiust tokensation during training

\subsection{Learning Tokenisation During Training}

Without being motivated by developmental plausibility, several works have explored learning to segment language inputs during training, for %during neural sequence modelling, 
tasks such as handwriting recognition \citep{kong-etal-2015-segmental}, %formal reasoning \citep{graves2017adaptive}, 
speech recognition \citep{wang-etal-2017-sequence}, and translation \citep{kreutzer-sokolov-2018-learning}. 
These efforts equip models with the ability to discover optimal segmentation units for a given task, rather than relying on a pre-determined segmentation scheme. 
%Both of these models use dynamic programming to efficiently compute the marginal likelihood during training and to find the optimal segmentation of a sequence.

%To achieve this for language modelling, 
\citet{sun-deng-2018-unsupervised} propose the segmental language model (SLM): an LSTM LM that computes the likelihood of a sentence by marginalising over all possible segmentations of the character sequence. Their aim is not improved performance, but rather unsupervised word segmentation for Chinese, which lacks explicit word boundaries. 
%Their Chinese SLM learns segmentations that are competitive with dedicated unsupervised word segmentation approaches. 
\citet{kawakami-etal-2019-learning} augment the SLM with a lexicon of high-frequency segments, improving unsupervised word segmentation.
%performance. % for Chinese and English (for the latter, they train on corpora with whitespaces removed).
\citet{downey-etal-2022-masked} propose the masked SLM, a bi-directional, transformer-based SLM that 
%model that marginalises over masked character spans, 
outperforms recurrent SLMs on unsupervised word discovery. % for Chinese and English.
SLMs are trained on raw character sequences without word boundaries (whitespaces) and, as a by-product of optimising the LM training objective, they partially recover word-level segmentation. 

\citet{meyer-buys-2022-subword} propose the subword segmental language model (SSLM), which adapts the SLM to model subword tokenisation. %, a foundational component of the modern LM pipeline.  
SSLM assumes access to word boundaries, constrains segments to subword units (they cannot cross word boundaries), and learns subword tokenisation to optimise the LM training objective. 
The original SSLM \citep{meyer-buys-2022-subword} is LSTM-based \citep{hochreiter-schmidhuber-1997-long}.
\citet{meyer-buys-2025-learning} propose a transformer \citep{vaswani-attention-2017} SSLM, primarily as a tool to study subword learning dynamics. 
For both models, evaluation is limited to the Nguni languages: low-resource, agglutinative languages, for which tokenisation was hypothesised to have an outsized impact. % on performance. 
SSLM outperformed tokenisation-based LMs on perplexity and sequence-to-sequence tasks, and performed strongly as an unsupervised morpheme segmenter.

It is unknown whether learnable subword tokenisation can improve pretraining sample-efficiency beyond the narrow linguistic scope of previous work. The BabyLM Challenge provides the ideal setting to test this. 
We propose two new SSLM variants, SubSegGPT and SubSegDeBERTa, which incorporate learnable subword tokenisation into modern, BabyLM-style architectures. % 
Understanding our models requires familiarity with the subword segmental framework, so in the next subsection we provide a technical overview of SSLM.

\subsection{Subword Segmental Language Modelling}
\label{subsec:background_subseg}

SSLM \citep{meyer-buys-2022-subword} establishes a framework 
%to unify language modelling and subword tokenisation in a single model that 
for learning tokenisation \emph{during} training.
% Any LM architecture can be adapted to the framework, 
% by modifying its training algorithm as follows. 
Any LM architecture can be adapted to the framework, removing its reliance on a fixed tokeniser. The key idea is that tokenisation is cast as a latent variable, inferred jointly with model parameters to optimise the LM objective. 
Adapting a model for subword segmental modelling requires augmenting its architecture with additional subnetworks, which we describe for our models in Section~\ref{sec:models}, and adopting the training algorithm of \citet{meyer-buys-2022-subword}, which we now summarise.

% The framework can be applied to any text generation task by modifying the standard autoregressive training algorithm and neural sequence modelling architecture to remove the reliance on a single, pre-determined subword tokenisation scheme. 

% The core idea behind subword segmental modelling is captured by its training algorithm. 

For a training sequence $S$, SSLM still minimises the standard LM loss $\mathcal{L}(\theta) = -\log p(S)$. However, whereas a vanilla LM computes $p(S)$ with the chain rule over a single pre-determined subword token sequence, % (produced during preprocessing by a tokeniser like BPE or ULM)
SSLM computes $p(S)$ as 
\begin{align} \label{marginalised_formula} 
p(S) = \sum_{T \in \pi(S)}p(T), 
\end{align}
where $\pi(S)$ is the set of all candidate tokenisations of $S$: every possible way that the sequence of words $S$ can be segmented into subword units (word boundaries are enforced, so subword units cannot span across whitespaces). Each $p(T)$ is still computed with the chain rule over the token sequence $T$, but the overall sequence probability $p(S)$ now incorporates multiple potential tokenisations of the training example $S$. 

Marginalising over $\pi(S)$ is intractable for long sequences, so SSLM introduces two constraints:
\begin{enumerate}
    \item Subword segments cannot exceed a maximum character length $L$, which is a hyperparameter.
    \item In computing the probability of a candidate tokenisation $T = \{t_1, t_2, ..., t_{|T|}\}$ with the chain rule, each next-token probability is conditioned on the \emph{untokenised} autoregressive (preceding) character-level context, so 
\begin{align} \label{tokenised_formula} 
p(T) = \prod_{i=1}^{|T|} p (t_i | t_{<i}) \approx \prod_{i=1}^{|T|} p (t_i | c_{<t_i}),
\end{align} 
 where $c_{<t_i}$ is the character sequence in $S$ that precedes $t_i$. This approximation discards tokenisation history, but enables tractable conditioning via a character-level context encoder. 
 %This encoder is the ``backbone'' of the subword segmental architecture: the deep neural network parameterising the LM probabilities of Equation~\ref{tokenised_formula}.
\end{enumerate}

Finally, to compute Equation~\ref{marginalised_formula} efficiently, \citet{meyer-buys-2022-subword} use a dynamic programming algorithm that iteratively computes $p(S_{1:k})$, the marginal probability of the sequence up to each character position $k$, for $k=1,..., |S|$ (we refer the reader to \citet{meyer-buys-2022-subword} for a detailed presentation of the algorithm).

% \begin{figure*}[t]
%     \centering
%     \begin{subfigure}[t]{0.49\textwidth}
%         \centering
%        \includegraphics[width=\linewidth]{figures/gpt.pdf}
%         \caption{Caption for the second figure.}
%         \label{fig:fig2}
%     \end{subfigure}
%     \begin{subfigure}[t]{0.49\textwidth}
%         \centering
%         \includegraphics[width=\linewidth]{figures/deberta.pdf}
%         \caption{Caption for the first figure.}
%         \label{fig:fig1}
%     \end{subfigure}
%     \hfill

%     \caption{Overall caption describing both figures.}
%     \label{fig:combined}
% \end{figure*}

\section{Models}
\label{sec:models}

The modelling assumptions and training algorithm outlined above describe the generative model of the subword segmental framework. Parameterising this with a neural architecture requires a model capable of computing $p (t_i | c_{<t_i})$ for \emph{any} subword token $t_i$ and a mechanism for conditioning this probability on the character-level context $c_{<t_i}$. A vanilla LM cannot assign probabilities to arbitrary subwords, so \citet{meyer-buys-2022-subword} augment the standard autoregressive architecture to enable this. 
Their methodology can be followed to adapt any architecture for subword segmental modelling. 
Doing so requires building an architecture with two components: an encoder that computes representations for a character-level context $c$ and a subword segment scorer that computes $p (t | c)$ for any subword $t$. 

In this section, we present SubSegGPT and SubSegDeBERTa, two new SSLMs that adapt respectively GPT-2 \citep{gpt2} and DeBERTa-v2 \citep{he2021deberta} for learnable subword tokenisation. Their architectures are visualised in Figure~\ref{fig:combined}.

\subsection{SubSegGPT}
\label{subsec:subseggpt}

SubSegGPT is a decoder-only SSLM with a GPT-2 backbone. It is a straightforward adaptation of the original, LSTM-based SSLM \citep{meyer-buys-2022-subword} to GPT-style language modelling and makes use of the same  training algorithm outlined in Section~\ref{subsec:background_subseg}.
We now describe its architecture, which closely mirrors the transformer-based SSLM of \citet{meyer-buys-2025-learning}, but is parameterised by the GPT-2 architecture to incorporate more recent architectural conventions and match the setup of competitive decoder-based BabyLMs.  

\subsubsection{Character-level history encoder} 
\label{subsubsec:subseggpt_encoder}
To compute $p(t_i | c_{<t_i})$ we need to encode $c_{<t_i}$, the full character sequence preceding the subword segment $t_i$. 
We encode $c_{<t_i}$ with a character-level GPT-2 backbone (excluding the language modelling head), using the final-layer output embedding $\mathbf{h}_{<t_i}$  of the last character before $t_i$ to represent the sequence history and to condition next-subword probabilities $p(t_i | \mathbf{h}_{<t_i})$.

\subsubsection{Subword segment scorer} 
\label{subsubsec:subseggpt_generator}
We follow previous SSLMs in computing $p(t_i | c_{<t_i})$ as a mixture of two subword probabilities, 
\begin{align} \label{mixture}
    p (t_i | c_{<t_i}) =\,\, & \lambda p_{\mathrm{lex}} (t_i | c_{<t_i})  + \nonumber\\
    & (1-\lambda) p_{\mathrm{char}} (t_i | c_{<t_i}),
\end{align}
where $\lambda \in (0,1)$ is a mixture coefficient, dynamically computed for each subword with a sigmoid-activated linear projection of $\mathbf{h}_{<t_i}$. 

$p_{\mathrm{lex}} (t_i | c_{<t_i})$ is a language modelling head that maps $\mathbf{h}_{<t_i}$ to a probability distribution over a fixed subword lexicon containing the $V$ most frequent subwords in the training corpus (the lexicon size $V$ is a pre-specified hyperparameter). $p_{\mathrm{char}} (t_i | c_{<t_i})$ is a small decoder subnetwork, parameterised by a 1-layer character-level LSTM,\footnote{SubSegGPT and SubSegDeBERTa are transformer-based LMs, parameterised by deep transformer backbones. LSTMs appear only as 1-layer subnetworks in the LM head, conditioning or generating subwords as short character sequences.}
that generates subword segment $t_i$ one character at a time and computes the subword probability as a chain rule product of individual character probabilities. It is conditioned on the sequence history by initialising the LSTM hidden state as $\mathbf{h}_{<t_i}$. 

The lexicon layer $p_{\mathrm{lex}}$ and character decoder $p_{\mathrm{char}}$ play complementary roles in next-subword prediction. 
$p_{\mathrm{lex}}$ directly computes probabilities of frequent subwords, such as common morphemes and words, but cannot handle infrequent, out-of-lexicon segments. 
By contrast, $p_{\mathrm{char}}$ can assign probabilities to arbitrary subword segments by composing them character by character, covering rare and previously unseen subwords. 
The mixture gate $\lambda$ learns to balance their contributions dynamically, based on the character-level context $\mathbf{h}_{<t_i}$.

The mixture-based subword segment scorer enables SubSegGPT to compute $p(t_i | c_{<t_i})$ for any candidate subword segment $t_i$ at any position in a training sequence. 
These probabilities (top of Figure~\ref{fig:combined}) are passed to the dynamic programming algorithm of Section~\ref{subsec:background_subseg}, which efficiently computes the marginal (Equation~\ref{marginalised_formula}) via the probabilities of all candidate tokenisations (Equation~\ref{tokenised_formula}). 
SubSegGPT is trained end-to-end by minimising the negative log-likelihood, jointly optimising autoregressive language modelling and subword tokenisation.

\subsection{SubSegDeBERTa}
\label{subsec:subsegdeberta}

We propose SubSegDeBERTa, an encoder-based, masked SSLM with a DeBERTa-v2 backbone. 
\citet{meyer-buys-2022-subword} introduce subword segmental modelling for decoder-only LMs. Their framework is inherently autoregressive, so its extension to masked language modelling is non-trivial. 
% The masked segmental LM of \citet{downey-etal-2022-masked} provides the closest precedent: a bi-directional transformer trained to jointly generate and segment masked spans. Their main application is unsupervised word discovery from small-scale corpora without word boundaries. 
% As a result, they make certain modelling decisions that are not suited to our aims. 
% First, to make up for small training sets, they mask and predict every possible segment of an input sequence in turn, unlike the random input masking (e.g. 15\% of token positions) of masked LM pretraining. 
% Second, they mask fixed-size continuous character segments in raw text sequences that lack explicit word boundaries. Their model learns unconstrained sequence segmentation, while we are focussed on the more constrained task of subword segmentation -- segments should not span across adjacent words and word boundary information should be encoded by the model.
In SubSegDeBERTa, we develop the first masked SSLM, incorporating learnable subword tokenisation into encoder-based pretraining. We encode bi-directional character-level context with a DeBERTa-v2 backbone, randomly mask a subset of input words, and jointly learn to generate and tokenise masked words into subword units.

% It mostly re-implements the transformer-based subword segmental LM of \citet{meyer-buys-2025-learning}, but is parameterised by the GPT-2 architecture to incorporate more modern, GPT-style conventions in decoder-based LM design. 
% SubSegDeBERTa is a novel, encoder-based architecture and the first instantiation of subword segmental modelling for masked language modelling. It jointly learns to generate and tokenise masked words during training, conditioned on bi-directional context from a DeBERTa-v2 backbone \citep{he2021deberta}.

\subsubsection{Character-level context encoder} 
\label{subsubsec:subsegdeberta_encoder}
We randomly mask a fixed proportion of words in each training sequence. 
We define words as whitespace-delimited character sequences and restrict subword segments to span within word boundaries. If a word is sampled for masking, we replace its entire character sequence with a single \texttt{[MASK]} token (as shown in Figure~\ref{fig:combined} for the masked word ``dogs''). 
Our bi-directional encoder is a character-level DeBERTa-v2 architecture that produces final-layer output embeddings for all characters, including \texttt{[MASK]} tokens. 
% We chose DeBERTa-v2 over other encoders because of its strong  performance as a backbone architecture in previous BabyLM submissions \citep{charpentier-etal-2025-findings}.

\subsubsection{Masked word scorer} 
\label{subsubsec:subsegdeberta_generator}
In vanilla MLMs, the final-layer representation $\textbf{h}_\mathrm{\texttt{[MASK]}}$ is used to predict the masked token. In SubSegDeBERTa, the masked word is not predicted as a single token. Instead, we generate masked words subword segmentally i.e. by marginalising over all possible subword tokenisations of a masked word. 

Suppose we mask the $i^\mathrm{th}$ word in a sentence, denoted by $w_i$ and consisting of characters $c_1, c_2, ..., c_{|w_i|}$. 
During training, we maximise 
\begin{align} \label{marginalised_word_formula} 
p(w_i|w_{<i}, w_{>i}) = \sum_{T \in \pi(w_i)}p(T |w_{<i}, w_{>i}), 
\end{align}
where $\pi(w_i)$ is the set of all candidate tokenisations of $w_i$, and word probabilities are conditioned on bi-directional context (all words to the left $w_{<i}$ and right $w_{>i}$ of the target word). 
The probability of each tokenisation $T = \{t_1, t_2, ..., t_{|T|}\}$ is computed with the chain rule as
\begin{align} \label{tokenised_word_formula} 
p(T |w_{<i}, w_{>i}) = \prod_{j=1}^{|T|} p (t_j | c_{<t_j}, w_{<i}, w_{>i}),
\end{align}
so this component of SubSegDeBERTa is autoregressive: masked word generation is conditioned on bi-directional context beyond the word ($w_{<i}, w_{>i}$), but within the word it is conditioned on left-to-right context ($c_{<t_j}$ is the character sequence preceding subword segment $t_j$ within word $w_i$).  

To compute the marginal of Equation~\ref{marginalised_word_formula} efficiently, we use the same dynamic programming algorithm discussed in Section~\ref{subsec:background_subseg} and introduce the same simplifying assumptions: we limit the length of $t_j$ to a pre-specified maximum number of characters and condition on untokenised character-level context, although here the context is bi-directional (Section~\ref{subsubsec:subsegdeberta_encoder}). 
To compute the subword probability $p (t_j | c_{<t_j}, w_{<i}, w_{>i})$, we use the same mixture model as SubSegGPT (Section~\ref{subsubsec:subseggpt_generator}). 
However, the masked LM setting introduces a complication that requires additional subnetworks.

The GPT-2 backbone of SubSegGPT outputs contextual representations for each character in an input sequence.
%, including characters that appear mid-word. 
These representations are passed to the subword segment scorer and are used to condition next-subword probabilities at any position in the character sequence (including subword segments that start mid-word, such as ``gs'' in ``dogs'', on the left of Figure~\ref{fig:combined}).  
The DeBERTa backbone of SubSegDeBERTa outputs contextual representations for all characters in a sequence \emph{except masked words}, whose characters are replaced by a single \texttt{[MASK]} token. 
%$\textbf{h}_\mathrm{\texttt{[MASK]}}$ encodes the bidirectional context $(w_{<i}, w_{>i})$, but the DeBERTa encoder cannot encode the autoregressive character-level context required to score individual subword segments in Equation~\ref{tokenised_word_formula}. 
% The DeBERTa backbone provides a single contextual representation $\textbf{h}_\mathrm{\texttt{[MASK]}}$ per masked word. 
% To compute $p(t_j | c_{<t_j}, w_{<i}, w_{>i})$ for all positions $j$ within $w_i$, we need a contextual representation for each character position in $w_i$ that encodes both the bidirectional context $(w_{<i}, w_{>i})$ and the autoregressive within-word history $c_{<t_j}$.
$\textbf{h}_\mathrm{\texttt{[MASK]}}$ encodes the bidirectional context $(w_{<i}, w_{>i})$, but provides only a single representation per masked word. To score subword segments at each position within $w_i$, we additionally need per-position representations that encode the autoregressive within-word history $c_{<t_j}$.

To address this, we introduce a \emph{word context encoder} to encode the context at every character in word $w_i$ (its role is visualised in Figure~\ref{fig:combined}).  %from $\textbf{h}_\mathrm{\texttt{[MASK]}}$. 
This is a 1-layer character-level LSTM that processes the target word's characters $c_1, c_2, ..., c_{|w_i|}$ left-to-right, conditioned on $\textbf{h}_\mathrm{\texttt{[MASK]}}$ in two ways: we initialise its hidden state as a learned projection of $\textbf{h}_\mathrm{\texttt{[MASK]}}$ and concatenate $\textbf{h}_\mathrm{\texttt{[MASK]}}$ to input character embeddings at every step. This produces a sequence of per-position context representations $\mathbf{h}_{k} = \mathrm{LSTM}(c_k, \textbf{h}_\mathrm{\texttt{[MASK]}})$,
% \begin{align} \label{word_context_formula}
% % \mathbf{h}_{c_{<t_j}} = \mathrm{LSTM}(c_{<t_j}, \textbf{h}_\mathrm{\texttt{[MASK]}}),
% \mathbf{h}_{k} = \mathrm{LSTM}(c_k, \textbf{h}_\mathrm{\texttt{[MASK]}}),
% \end{align}
that encode the bidirectional context $(w_{<i}, w_{>i})$ via $\textbf{h}_\mathrm{\texttt{[MASK]}}$ and the within-word history ($c_{<t_j}$) via LSTM recurrence. This provides a contextual representation $\mathbf{h}_{k}$ for every character position in $w_i$, which is used to condition $p(t_j | c_{<t_j}, w_{<i}, w_{>i})$ for any subword segment $t_j$ in $w_i$, as shown for all possible subwords in the word ``dogs'', on the right of Figure~\ref{fig:combined}.

To compute these probabilities, we use the same mixture model as the SubSegGPT subword segment scorer (Equation~\ref{mixture}), with $\mathbf{h}_{k}$ passed as conditioning context, instead of $\mathbf{h}_{<t_i}$. The dynamic programming algorithm of Section~\ref{subsec:background_subseg} computes the masked word marginal (Equation~\ref{marginalised_word_formula}) via the probabilities of all candidate tokenisations (Equation~\ref{tokenised_word_formula}). SubSegDeBERTa is trained end-to-end by minimising the negative log-likelihood of masked words, jointly optimising masked language modelling and subword tokenisation.

\begin{table*}[t!]
\centering
\small
\begin{tabular}{llccccccc}
\toprule
\textbf{Track} & \textbf{Model} & \textbf{BLiMP} & \textbf{BLiMP Sup.} & \textbf{EWoK} & \textbf{Entity} & \textbf{COMPS} & \textbf{PIQA} & \textbf{Avg.} \\
\midrule
& Random chance & 50.00 & 50.00 & 50.00 & 20.00 & 50.00 & 37.50 & 42.92 \\
\midrule
\multirow{4}{*}{\textsc{Strict}} & GPT-2	&74.73	&65.00 &	54.37	&16.91	&55.85	&36.62&	50.58 \\
&SubSegGPT&	\textbf{78.22}	&64.78&	51.72&	16.12&	53.75&	\textbf{40.15}			& 50.79						\\
&DeBERTa	&73.95&	64.20	&52.35	&17.46	&53.13&	35.30	&49.40						\\
% &SubSegDeBERTa&	\textbf{78.90}	&\textbf{67.65}&	\textbf{56.94}	&\textbf{20.58}	&\textbf{58.16}&	\textbf{38.56}&	\textbf{53.47}\\
 &SubSegDeBERTa&
\textbf{78.22}&
\textbf{68.73}&
\textbf{55.39}&
\textbf{22.28}&
\textbf{57.77}&
40.02&
\textbf{53.74}\\
\midrule
\multirow{4}{*}{\textsc{Strict-Small}} &GPT-2	&65.23&	57.25&	50.63	&19.1&	\textbf{51.81}	&35.09&	46.52\\
&SubSegGPT&	\textbf{68.80}	&\textbf{62.77}&	49.76&	\textbf{20.48}	&50.51&	30.67&	\textbf{47.17}\\
&DeBERTa&	63.78	&59.94&	50.28	&19.67&	50.88&	\textbf{35.28}	&46.64						 \\
&SubSegDeBERTa	&64.68	&60.56&	\textbf{50.99}	&17.89	&51.28	&33.74&	46.52 \\
\bottomrule
\end{tabular}
\caption{Results for zero-shot tasks in the BabyLM evaluation pipeline. Best result per track is \textbf{boldfaced}.}
\vspace{-0.25cm}
\label{tab:results_zeroshot}
\end{table*}

\section{Experimental Setup}
\label{sec:experimental_setup}

\subsection{Pretraining}

We follow the guidelines of the 2026 BabyLM Challenge \citep{choshen2026babylmturns4goes} to pretrain SubSegGPT and SubSegDeBERTa for \textsc{Strict} and \textsc{Strict-small}. 
We use the text-only datasets released by the organisers and pretrain for 10 epochs. %, consisting of respectively 100M and 10M words of detoxified, developmentally plausible English text. 
%For the \textsc{Multilingual} track, contestants are encouraged to experiment with their own mixtures of English, Dutch, and Chinese data from \textsc{BabyBabelLM} \citep{jumelet-etal-2026-babybabellm}, constrained to the equivalent of 100M English words, adjusted for the byte premium of each language \citep{arnett-etal-2024-bit}. 
%Our research does not explore training data curation or language ratio, so we simply use the training mixture used by the BabyLM Challenge \textsc{Multilingual} baselines, which is created by naively sampling an equal number of byte premium adjusted words from the English, Dutch, and Chinese \textsc{BabyBabelLM} datasets. 
% Our models are pretrained on text-only and we do not explore multimodal or interaction-based pretraining.

To test the impact of learnable tokenisation, we compare our models to fixed-tokenisation BabyLMs based on the corresponding backbone architectures and trained on the same data. 
For SubSegGPT, we use the GPT-2 \citep{gpt2} baselines released by the BabyLM Challenge organisers. %\footnote{\href{https://huggingface.co/collections/BabyLM-community/babylm-2026-baselines}{\tt babylm-2026-baselines}}
For SubSegDeBERTa, we pretrain our own DeBERTa-v2 \citep{he2021deberta} models as baselines,
%\footnote{\href{https://huggingface.co/collections/francois-meyer/deberta-babylm-baselines}{\tt deberta-babylm-baselines}} 
matching the backbone architectural configurations of SubSegDeBERTa for comparability. Table~\ref{tab:backbone_hyperparameters} in the appendix details our baseline setup.
%We pretrain each of our models on a single A100 GPU. 

%                 Strict          Strict-small
% GPT-2           
% SubSegGPT       85h             16h

% DeBERTa         17h             2h
% SubSegDeBERTa   80h             9h

\subsection{Hyperparameters}

%We pretrain each model for 10 epochs, 

%As required by the official evaluation pipeline, we save intermediate model checkpoints. 
% at regular intervals. % (every 1M words until 10M words, every 10M words until 100M words, and every 100M words until 1B words).
Our backbone architectures match the size of the BabyLM GPT-2 baseline, which corresponds to \textsc{Base} configurations of GPT-2 and DeBERTa. 
%(12 layers, 12 attention heads, 768-dimensional embeddings, 3072-dimensional hidden layers, and a sequence length of 1,024).
We tune pretraining hyperparameters as detailed in Appendix~\ref{appendix:hyperparams}. % in the appendix. 
Our backbone architectures are character-based, so their embedding matrices contribute negligibly to overall parameter count. 
However, our subnetworks for subword segment scoring introduce additional parameters. 
The model sizes and hyperparameter settings of our  submissions and baselines are reported in Tables~\ref{tab:backbone_hyperparameters} and~\ref{tab:segmental_hyperparameters} in the appendix.

% batch_size
% warmup_ratio
% mlm_probability
% max_seg_len
% lex_vocab_size

\subsection{Evaluation}
We evaluate on the official 2026 BabyLM Challenge evaluation pipeline,
%for \textsc{Strict} and \textsc{Strict-small}, 
which includes three types of tasks (full list in Appendix~\ref{appendix:evaluation}).
\textbf{(1) Zero-shot tasks} test linguistic knowledge directly from model probabilities %, without task-specific finetuning. 
via minimal pair evaluations. % in the style of BLiMP \citep{warstadt-etal-2020-blimp}. 
%It also includes Global PIQA \citep{mrl-workshop-2025-global-piqa}, a commonsense reasoning task announced shortly before the competition deadline.
%For the \textsc{Multilingual} track, this includes the zero-shot tasks for English, Dutch, and Chinese from the BabyBabelLM evaluation benchmark \citep{jumelet-etal-2026-babybabellm}.
\textbf{(2) Finetuning tasks} test natural language understanding on (Super)GLUE \citep{wang-etal-2018-glue, superglue} with task-specific finetuning.  %For the \textsc{Multilingual} track, this includes finetuning tasks for English, Dutch, and Chinese from the BabyBabelLM evaluation benchmark \citep{jumelet-etal-2026-babybabellm}.
\textbf{(3) Human-likeness tasks} evaluate how well model predictions align with human psycholinguistic data (we report these results in Appendix \ref{appendix:human_likeness_results}, as neither our models nor baselines perform well on these tasks). %These tasks 
%are included in the \textsc{Strict} and \textsc{Strict-small} tracks only, and 
% consist of reading time correlations \citep{devarda2024cloze} and age-of-acquisition scores \citep{chang-bergen-2022-word}. % based on intermediate pretraining checkpoints. 

%CLAude drafted

\subsection{Evaluating Subword Segmental LMs}

The BabyLM evaluation pipeline is designed for standard LM architectures. 
To evaluate our models on finetuning tasks, we can apply the pipeline without change (final-layer representations are passed to classification heads).
However, for zero-shot tasks, the pipeline expects per-position logits, which our models do not emit: SubSegGPT outputs a per-sentence marginal (Equation~\ref{marginalised_formula}) and SubSegDeBERTa outputs a per-word marginal (Equation~\ref{marginalised_word_formula}). 
%Zero-shot and human-likeness tasks depend on per-position log-probabilities, so
We implement model-specific wrappers that transform our outputs into the quantities required for each task. %: per-character log-probabilities for SubSegGPT and per-word cloze probabilities for SubSegDeBERTa. 
We leave the official pipeline unchanged, except for one line of code (see Appendix ~\ref{appendix:evaluating_subseggpt} for details). Appendix~\ref{appendix:evaluating_our_models} describes the wrappers we implement to evaluate our models.

% The finetuning evaluation pipeline can be applied directly, since it does not rely on per-position logits: it passes final-layer hidden representations from our model backbones to a task-specific classification head.
%(for sequence classification tasks, SubSegDeBERTa uses the \texttt{[CLS]} representaion and SubSegGPT uses the representation of the final character in the sequence). 

\begin{table*}[t!]
\centering
\small
\begin{tabular}{llcccccccc}
\toprule
\textbf{Track} &\textbf{Model} & \textbf{BoolQ} & \textbf{MNLI} & \textbf{MRPC} & \textbf{MultiRC} & \textbf{QQP} & \textbf{RTE} & \textbf{WSC} & \textbf{Avg.}\\
\midrule
& Majority class & 64.04 & 35.70 & 68.14 & 57.55 & 62.78 & 53.96 & 61.54 & 57.67 \\
\midrule
\multirow{4}{*}{\textsc{Strict}} &GPT-2&	69.66&	60.76&	85.34&	65.92&	71.56	&57.55	&63.46&	67.75\\
&SubSegGPT	&64.59&	58.21&	84.51	&66.38	&72.34	&60.43&	63.46	&67.13				\\
&DeBERTa	&\textbf{72.66}	&\textbf{62.59}	&84.80	&\textbf{68.15}&	\textbf{78.64}	&\textbf{66.19}&	63.46&	\textbf{70.93}							\\
%&SubSegDeBERTa	&72.42&	60.94	&\textbf{88.97}	&\textbf{66.83}&	75.36	&61.15&	63.46&	69.88\\
&SubSegDeBERTa&
71.44  &
60.80  &
\textbf{90.54}  &
67.16&
75.16&
64.03&
63.46&
70.37 \\
\midrule
\multirow{4}{*}{\textsc{Strict-Small}} &GPT-2	&67.71	&49.84&	81.37	&65.76	&61.67	&56.83	&63.46	&63.81\\
&SubSegGPT&	64.83	&\textbf{54.65}	&\textbf{85.71}	&65.64&	\textbf{71.23}&	\textbf{60.43}&	63.46&	\textbf{66.56}\\
&DeBERTa	&67.52	&47.29	&70.59	&\textbf{67.53}	&70.17	&56.83&	61.54	&63.07 	\\
&SubSegDeBERTa&	\textbf{68.93}	&45.78&	81.93&	65.80	&62.94&	56.83&	\textbf{65.38}&	63.94\\
\bottomrule
\end{tabular}
\caption{Results for finetuning tasks in the BabyLM evaluation pipeline. Best result per track is \textbf{boldfaced}.}
\label{tab:results_finetuning}
\vspace{-0.25cm}
\end{table*}

\section{Results}
\label{sec:results}

% \paragraph{Zero-shot}

Table~\ref{tab:results_zeroshot} reports zero-shot results. In the \textsc{Strict} track, SubSegDeBERTa achieves the highest scores across most tasks, comfortably outperforming the strongest baseline, GPT-2. SubSegGPT also outperforms both tokenisation-based baselines, suggesting that at this scale (10 epochs over 100M words) learning tokenisation during pretraining reliably improves sample-efficiency of intrinsic linguistic knowledge acquisition.

In the \textsc{Strict-small} track, the relative performances of our two models are reversed. 
SubSegGPT outperforms SubSegDeBERTa and, on average, outperforms both baselines. 
Performance is more mixed across individual tasks -- largely because, for most tasks, all models fail to reach above-chance performance. On the only two tasks where models reliably outperform chance, SubSegGPT achieves large performance gains over its tokenisation-based equivalent, GPT-2 (+3.57 for BLiMP and +5.52 for BLiMP Supplement). SubSegDeBERTa does not outperform DeBERTa in the \textsc{Strict-small} track, but reaches similar performance levels on average.

Overall, these results suggest that the benefits of learnable tokenisation are scale-dependent. In the extremely low-resource setting, where even above-chance zero-shot performance is challenging, SubSegGPT already offers gains. 
SubSegDeBERTa requires a larger data scale for its sample-efficiency to take effect, but under those conditions it produces even more reliable gains than SubSegGPT. The relative underperformance of SubSegDeBERTa in the \textsc{Strict-small} track might be due to its MLM objective, which provides a sparser training signal than autoregressive modelling. %\citep{Clark2020ELECTRA}.
SubSegDeBERTa is only trained to generate a proportion of words in each sequence, whereas SubSegGPT is trained to generate every word in the corpus. 
% Given enough data to offset this sparser signal, however, SubSegDeBERTa becomes the most sample-efficient model of all, as its \textsc{Strict} results show.

% \paragraph{Finetuning}

Table~\ref{tab:results_finetuning} reports results for text classification and entailment tasks from the (Super)GLUE benchmark. With task-specific finetuning, the benefits of subword segmental pretraining diminish. In the \textsc{Strict} track, neither of our models consistently outperform their baselines. 
Among all the models we tested, DeBERTa achieves the highest average performance. At the scale of 100M words, conventional encoder-only pretraining with fixed subword tokenisation, combined with downstream finetuning, is sufficient for natural language understanding tasks. 
In the \textsc{Strict-small} track, SubSegGPT again achieves the best performance overall, as it did in zero-shot evaluation. This supports our claim that SubSegGPT provides better sample-efficiency when pretraining data is severely limited, and that this pretraining sample-efficiency transfers to improved downstream finetuning. 
% You lose subword segmentality

\begin{figure*}[t!]
    \centering
    \includegraphics[width=\linewidth]{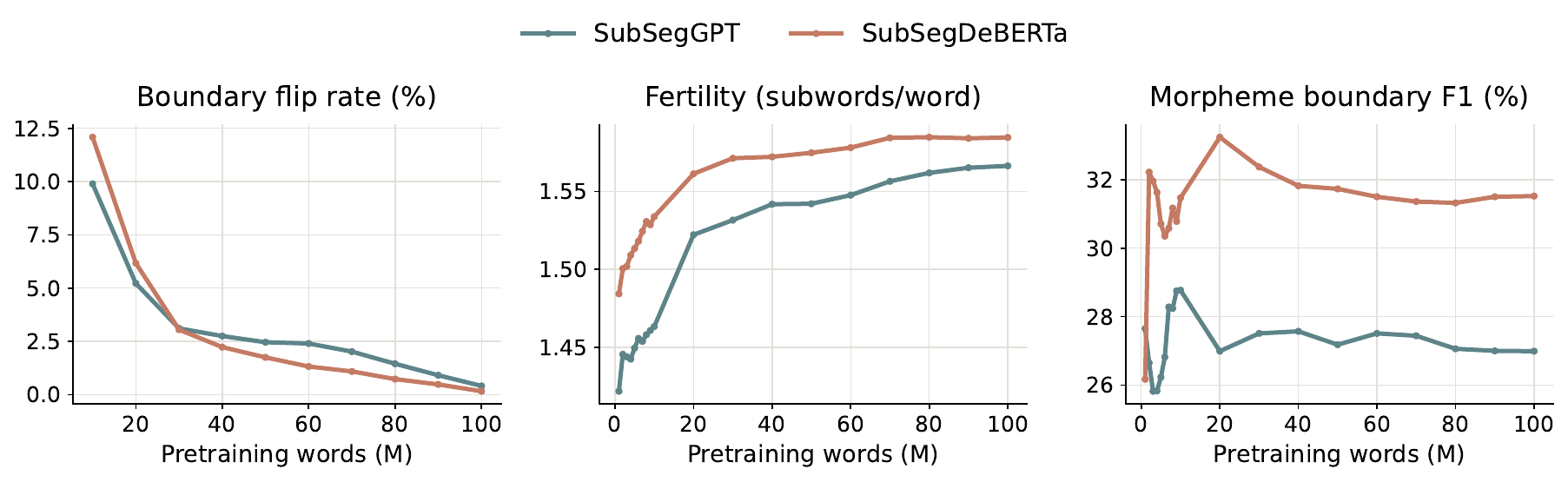} % Scales the PDF and selects page 1
   %\vspace{-0.5cm}
    \caption{Subword learning dynamics of SubSegGPT and SubSegDeBERTa during \textsc{Strict-small} pretraining. }
    \label{fig:my_pdf_figure}
\end{figure*}

\newcommand{\morphemegreen}[1]{\tikz[baseline]{\node[draw=green!50!black, fill=green!10, rounded corners=2pt, inner sep=2pt, outer sep=0pt, anchor=base]{\strut #1};}}
\newcommand{\morphemeplain}[1]{%
  \tikz[baseline]{\node[draw=gray, rounded corners=2pt, inner sep=2pt, outer sep=0pt, anchor=base]{\strut #1};}%
}

\begin{table*}[t] \small
    \centering
    \renewcommand{\arraystretch}{1.3}
    \setlength{\tabcolsep}{4pt}
    \begin{tabular}{l|l|l}
    \toprule
        & \textbf{SubSegGPT} & \textbf{SubSegDeBERTa} \\
    \midrule
    \textbf{1M} & \morphemegreen{yes} \morphemegreen{he}\morphemeplain{\textquotesingle{}}\morphemegreen{s} \morphemeplain{playi}\morphemeplain{ng} \morphemegreen{in} \morphemegreen{the} \morphemegreen{sand}\morphemegreen{box}\morphemeplain{.} & \morphemegreen{yes} \morphemegreen{he}\morphemeplain{\textquotesingle{}}\morphemegreen{s} \morphemegreen{play}\morphemegreen{ing} \morphemegreen{in} \morphemegreen{the} \morphemeplain{s}\morphemeplain{and}\morphemegreen{box}\morphemeplain{.} \\
    \textbf{10M} & \morphemegreen{yes} \morphemegreen{he}\morphemeplain{\textquotesingle{}}\morphemegreen{s} \morphemeplain{playi}\morphemeplain{ng} \morphemegreen{in} \morphemegreen{the} \morphemegreen{sand}\morphemegreen{box}\morphemeplain{.} & \morphemegreen{yes} \morphemegreen{he}\morphemeplain{\textquotesingle{}}\morphemegreen{s} \morphemegreen{play}\morphemegreen{ing} \morphemegreen{in} \morphemegreen{the} \morphemegreen{sand}\morphemegreen{box}\morphemeplain{.} \\
    \textbf{100M} & \morphemegreen{yes} \morphemegreen{he}\morphemeplain{\textquotesingle{}}\morphemegreen{s} \morphemeplain{playi}\morphemeplain{ng} \morphemegreen{in} \morphemegreen{the} \morphemegreen{sand}\morphemegreen{box}\morphemeplain{.} & \morphemegreen{yes} \morphemegreen{he}\morphemeplain{\textquotesingle{}}\morphemegreen{s} \morphemegreen{play}\morphemegreen{ing} \morphemegreen{in} \morphemegreen{the} \morphemegreen{sand}\morphemegreen{box}\morphemeplain{.} \\
    \midrule
    \textbf{1M} & \morphemegreen{six} \morphemeplain{littl}\morphemeplain{e} \morphemeplain{ta}\morphemeplain{xis} \morphemegreen{stand}\morphemegreen{ing} \morphemegreen{in} \morphemegreen{a} \morphemegreen{row}\morphemeplain{.} & \morphemegreen{six} \morphemeplain{l}\morphemeplain{ittle} \morphemeplain{ta}\morphemeplain{xis} \morphemegreen{stand}\morphemegreen{ing} \morphemegreen{in} \morphemegreen{a} \morphemegreen{row}\morphemeplain{.} \\
    \textbf{10M} & \morphemegreen{six} \morphemeplain{littl}\morphemeplain{e} \morphemeplain{ta}\morphemeplain{xi}\morphemegreen{s} \morphemegreen{stand}\morphemegreen{ing} \morphemegreen{in} \morphemegreen{a} \morphemegreen{row}\morphemeplain{.} & \morphemegreen{six} \morphemeplain{littl}\morphemeplain{e} \morphemeplain{t}\morphemeplain{ax}\morphemeplain{i}\morphemegreen{s} \morphemegreen{stand}\morphemegreen{ing} \morphemegreen{in} \morphemegreen{a} \morphemegreen{row}\morphemeplain{.} \\
    \textbf{100M} & \morphemegreen{six} \morphemeplain{littl}\morphemeplain{e} \morphemeplain{ta}\morphemeplain{xi}\morphemegreen{s} \morphemeplain{s}\morphemeplain{tandi}\morphemeplain{ng} \morphemegreen{in} \morphemegreen{a} \morphemegreen{row}\morphemeplain{.} & \morphemegreen{six} \morphemeplain{littl}\morphemeplain{e} \morphemeplain{ta}\morphemeplain{xi}\morphemegreen{s} \morphemegreen{stand}\morphemegreen{ing} \morphemegreen{in} \morphemegreen{a} \morphemegreen{row}\morphemeplain{.} \\
    \midrule
    \textbf{1M} & \morphemegreen{you}\morphemeplain{\textquotesingle{}}\morphemegreen{re} \morphemeplain{jumpi}\morphemeplain{ng} \morphemegreen{in} \morphemegreen{Mommy}\morphemeplain{\textquotesingle{}}\morphemegreen{s} \morphemeplain{kitch}\morphemeplain{en}\morphemeplain{?} & \morphemegreen{you}\morphemeplain{\textquotesingle{}}\morphemegreen{re} \morphemegreen{jump}\morphemegreen{ing} \morphemegreen{in} \morphemegreen{Mommy}\morphemeplain{\textquotesingle{}}\morphemegreen{s} \morphemeplain{kitch}\morphemeplain{en}\morphemeplain{?} \\
    \textbf{10M} & \morphemegreen{you}\morphemeplain{\textquotesingle{}}\morphemegreen{re} \morphemeplain{jumpi}\morphemeplain{ng} \morphemegreen{in} \morphemegreen{Mommy}\morphemeplain{\textquotesingle{}}\morphemegreen{s} \morphemeplain{ki}\morphemeplain{tchen}\morphemeplain{?} & \morphemegreen{you}\morphemeplain{\textquotesingle{}}\morphemegreen{re} \morphemegreen{jump}\morphemegreen{ing} \morphemegreen{in} \morphemegreen{Mommy}\morphemeplain{\textquotesingle{}}\morphemegreen{s} \morphemeplain{kitch}\morphemeplain{en}\morphemeplain{?} \\
    \textbf{100M} & \morphemegreen{you}\morphemeplain{\textquotesingle{}}\morphemegreen{re} \morphemeplain{jumpi}\morphemeplain{ng} \morphemegreen{in} \morphemegreen{Mommy}\morphemeplain{\textquotesingle{}}\morphemegreen{s} \morphemeplain{ki}\morphemeplain{tchen}\morphemeplain{?} & \morphemegreen{you}\morphemeplain{\textquotesingle{}}\morphemegreen{re} \morphemegreen{jump}\morphemegreen{ing} \morphemegreen{in} \morphemegreen{Mommy}\morphemeplain{\textquotesingle{}}\morphemegreen{s} \morphemeplain{ki}\morphemeplain{tchen}\morphemeplain{?} \\
    
    \bottomrule
    \end{tabular}
    \caption{Learned subword tokenisations of CHILDES utterances across \textsc{Strict-small} pretraining checkpoints, showing how subword boundaries evolve and highlighting subwords that correspond to morphemes.}
    \label{table_childes_examples_1}
    \vspace{-0.25cm}
\end{table*}

\section{Analysing Subword Learning}

In SubSegGPT and SubSegDeBERTa, we can study subword tokenisation as a learnable component of language modelling: equipped with the ability to learn tokenisation, how do subword boundaries evolve over pretraining and what are the linguistic properties of the final subword units?

\subsection{Subword Learning Dynamics}

% SubSegGPT and SubSegDeBERTa dynamically optimise subword tokenisation during pretraining. 
% Unlike for vanilla LMs with fixed tokenisers, we can study the subword learning dynamics of language modelling: how tokenisation evolves over pretraining checkpoints.
%Unlike with vanilla LMs, which are based on fixed tokenisers, 
%We can study  %\citep{meyer-buys-2025-learning}
We compare tokenisations of the 2022 SIGMORPHON English test set \citep{batsuren-etal-2022-sigmorphon} across regular interval checkpoints (every 1M words until 10M words, every 10M words until 100M words).
To extract the learned tokenisation of a sentence, we use the Viterbi algorithm to extract the highest-probability tokenisation (replacing the sum in Equations~\ref{marginalised_formula} and~\ref{marginalised_word_formula} with an argmax).

For each checkpoint, we quantify the properties of its subwords with three metrics. 
(1) Boundary flip rate measures the rate of change in tokenisation as the fraction of possible subword boundaries that change from one checkpoint to the next. 
(2) Fertility \citep{acs-2021-exploring} is the average number of subwords per word, reflecting the granularity of subword tokenisation. 
(3) Morpheme boundary identification F1 measures the overlap between learned subword boundaries and ground truth morpheme boundaries, as annotated in the SIGMORPHON dataset.

Figure~\ref{fig:my_pdf_figure} plots the learning dynamics of SubSegGPT and SubSegDeBERTa during \textsc{Strict-small} pretraining. 
The two models exhibit similar learning trajectories. 
After an initial period of rapid changes, subword learning gradually stabilises and converges on a settled tokenisation scheme. 
This convergence is characterised by a steady increase in fertility: words are tokenised into more subword units. As a comparison, the fertility of the BabyLM baseline tokeniser (16k-vocabulary BPE) on this dataset is 1.23, so our models learn more aggressive segmentation.
The subword boundaries of both models shift towards greater alignment with morphological boundaries. This alignment remains weak compared to a dedicated unsupervised morphological segmenter like Morfessor \citep{smit-etal-2014-morfessor}, which achieves 42.9\% F1 on the same evaluation set, but is well above random boundary insertion at the same fertility as our models, which would achieve around 11.6\% F1. 

\subsection{Qualitative Analysis}

The subword units learned by SubSegGPT and SubSegDeBERTa reflect the tokenisation demands of sample-efficient language modelling. 
Our quantitative analysis suggests that this balances linguistic plausibility (partial morphological alignment) against frequency-based criteria (finer-grained segmentation).
%The subword learning process is also biased by our subword segmental architecture design (e.g. our mixture model for subword generation) and constrained by our modelling assumptions (e.g. we limit subword segments to maximum 5 characters). 
To study this trade-off qualitatively, Table~\ref{table_childes_examples_1} presents CHILDES \citep{macwhinney2000childes} utterances tokenised by our models, highlighting subwords corresponding to morphemes.

The examples show several instances of our models discovering morphemes as subword units, such as suffixes (``--ing'', ``--s'') and compound constituents (``sand--box'').
They also show instances of morphologically unsound tokenisation, some of which are model-specific: SubSegGPT performs worse as a morphological segmenter  (see Figure~\ref{fig:my_pdf_figure}) and this is reflected in examples like ``playi--ng'' and ``jumpi--ng''. %, treating ``--ng'' rather than ``--ing'' as the suffix. This systematic error persists across checkpoints and is one source of its . 
Other morphological violations can be attributed to biases introduced by our architecture design (e.g. our mixture model for subword generation) and constraints imposed by modelling assumptions. For example, subword segments cannot exceed 5 characters, so words like ``kitchen'' and ``little'' cannot be left untokenised.
Removing these constraints would allow us to study truly unrestricted subword learning, but marginalising over unbounded segment lengths is computationally infeasible in our current setup.

% BPE:  subwords/word, morpheme boundary F1 = 13.4%

%                     Morfessor2	  SubSegGPT	      SubSegDeBERTa 	   BPE (16k)
% Morpheme  F1	    42.9	      27.0            31.5                 13.4
% Fertility	        1.49	      1.57	          1.58	               1.23

% \begin{figure*}[htbp]
%     \centering
%     \includegraphics[width=\linewidth]{figures/row_smooth-dots.pdf} % Scales the PDF and selects page 1
%     \caption{This figure was imported directly from a PDF file.}
%     \label{fig:my_pdf_figure}
% \end{figure*}

\label{sec:analysis}

\section{Conclusion}

%This paper investigates whether learning subword tokenisation during pretraining improves sample efficiency. 
%We 
We present two new models that learn subword tokenisation during training to optimise their pretraining objectives: SubSegGPT for autoregressive language modelling and SubSegDeBERTa for masked language modelling. 
% Our results show that learnable subword tokenisation, which we argue is more developmentally plausible than fixed tokenisation, can improve performance under constrained pretraining budgets. 
Our models excel at different scales:
SubSegDeBERTa is our most competitive submission in the \textsc{Strict} track and SubSegGPT performs strongly in the \textsc{Strict-small} track. 
By combining the SSLM framework with current best practices in sample-efficient pretraining, we show that SSLMs are effective beyond the low-resource, agglutinative languages for which they were originally proposed.
More generally, our findings suggest reconsidering tokenisation conventions and identify end-to-end subword modelling as a promising future direction for BabyLM research.
%More generally, our work emphasizes that tokenisation is something worth investigating as it could be playing arole on sample efficiency and.
% submission to the 2026 BabyLM Challenge
% Both of these models learn subword boundaries during training, allowing them to opptimise subword tokenisation for developmentallyt plausible training. We argue that this is more aligned with how humans learn language -- it's more developmentally plausible -- than fixed subword tokenisation.
% Our results show that learnable subword tokenisation, which we argue is more developmentally plausible than fixed tokenisation, is also more sample-efficient for BabyLM pretraining. 

\label{sec:conclusion}

\section{Limitations}

A disadvantage of subword segmental modelling is the additional computational complexity introduced by its training algorithm. Marginalising over several tokenisations requires more computations than using a single tokenisation, so SubSegGPT and SubSegDeBERTa have much longer training times than fixed-tokenisation BabyLMs (A100 GPU hours are listed in Table~\ref{tab:backbone_hyperparameters} in the appendix). 
For example, training SubSegDeBERTa required approximately 5$\times$ the A100 GPU hours of DeBERTa.
Our aim is sample-efficiency (better performance with the same pretraining data budget), but this comes at the cost of compute-efficiency (more training FLOPs and longer training times). 
%For example, training SubSegDeBERTa required approximately five times as many A100 GPU hours as training the corresponding DeBERTa models.

% \include{tables/performance_summary}

% ---- Bibliography inlined for arXiv (was: \bibliography{custom, anthology-1, anthology-2}) ----

% \appendix
\appendix
\section{ Hyperparameters}
\label{appendix:hyperparams}
During development, we tuned pretraining hyperparameters by comparing performance on the \textsc{Fast} evaluation pipeline, which samples subsets of the zero-shot BabyLM evaluation datasets. 
Among standard pretraining hyperparameters, we tuned the learning rate, batch size, warmup ratio, and masking ratio (for SubSegDeBERTa). Among hyperparameters unique to subword segmental modelling, we tuned the maximum subword segment length and the subword lexicon size (the number of top-frequency character n-grams to include in the lexicon subword scorer).  

We did not perform a full grid search. Instead, we started with the BabyLM baseline hyperparameters as our default setup and varied one hyperparameter at a time. This revealed that changing certain hyperparameters (batch size, warmup ratio, and maximum segment length beyond 5 characters) had little effect on performance, while others (learning rate, masking ratio, and subword lexicon size) were influential. We subsequently experimented with different learning rates (1e-3, 1e-4, 5e-4), masking ratios (0.15, 0.3, 0.4, 0.5), and subword lexicon sizes (5k, 10k, 20k, 40k), conducting most experiments in the \textsc{Strict-small} setting due to its faster pretraining iterations, with more limited experimentation in the \textsc{Strict} setting. The hyperparameters of our final submissions are reported in Tables~\ref{tab:backbone_hyperparameters} and~\ref{tab:segmental_hyperparameters}. 

\begin{table*}[t]
  \centering
  \small
  \setlength{\tabcolsep}{5pt}
  \begin{tabular}{l cc cc}
    \toprule
    %  & \multicolumn{2}{c}{\textsc{GPT-2 family}} & \multicolumn{2}{c}{\textsc{DeBERTa-v2 family}} \\
    % \cmidrule(lr){2-3}\cmidrule(lr){4-5}
    \textbf{Hyperparameter} & \textbf{GPT-2} & \textbf{SubSegGPT} & \textbf{DeBERTa-v2} & \textbf{SubSegDeBERTa} \\
    \midrule
    % \multicolumn{5}{l}{\textit{Architecture}} \\
     & \textsc{Strict}\,/\,\textsc{Small}& \textsc{Strict}\,/\,\textsc{Small}& \textsc{Strict}\,/\,\textsc{Small}& \textsc{Strict}\,/\,\textsc{Small} \\
     \midrule
    A100 training  & --      & 85h\,/\,16h & 17h\,/\,2h & 80h\,/\,9h \\
    Parameters     & 98.4M      & 95.8\,/\,103.1M & 113.2M & 140.4\,/\,116.9M \\
    Layers         & 12         & 12         & 12         & 12         \\
    Hidden size              & 768        & 768        & 768        & 768        \\
    FF size     & 3,072     & 3,072     & 3,072     & 3,072     \\
    Attention heads          & 12         & 12         & 12         & 12         \\
    Dropout           & 0.1        & 0.1        & 0.1        & 0.1        \\
    Vocabulary size          & 16,384    & --    & 16,384    & --     \\
    Sequence length    & 512        & 1,024     & 512        & 1,024     \\
    % \midrule
    % \multicolumn{5}{l}{\textit{Optimisation}}\\
    Learning rate            & 5e-5       & 5e-4       & 5e-5       & 5e-4       \\
    LR scheduler             & cosine     & cosine     & cosine     & cosine     \\
    Warmup ratio             & 0.01       & 0.1        & 0.01       & 0.1        \\
    Weight decay             & 0          & 0.01       & 0          & 0.01       \\
    Gradient clipping        & 1.0        & 1.0        & 1.0        & 1.0        \\
    Masking ratio        & --         & --         & 0.3        & 0.3\,/\,0.4        \\
    Batch size    & 16         & 16         & 16         & 16         \\
    \bottomrule
  \end{tabular}
  \caption{Backbone architecture configurations and training hyperparameters for our submissions and baselines.}
  \label{tab:backbone_hyperparameters}
\end{table*}

% ------------------------------------------------------------------
% TABLE 2: Subword-segmental-specific components
% ------------------------------------------------------------------
\begin{table}[t]
  \centering
  \small
  \setlength{\tabcolsep}{5pt}
  \begin{tabular}{l cc cc}
    \toprule
     & \multicolumn{2}{c}{\textbf{SubSegGPT}} & \multicolumn{2}{c}{\textbf{SubSegDeBERTa}} \\
    \cmidrule(lr){2-3}\cmidrule(lr){4-5}
     & \textsc{Strict} & \textsc{Small} & \textsc{Strict} & \textsc{Small} \\
    \midrule
    %Parameters            & 95.8M & 103.1M & 140.4M & 116.9M \\
    Char vocab   & 742   & 416    & 744    & 418    \\
    %\quad of which alphabetic ($n_\alpha$) & 426 & 263 & 426 & 263 \\
    Lexicon size & 10k & 20k & 40k & 10k \\
    %Lexicon min.\ frequency     & 5     & 5      & 5      & 5      \\
    Max segment & 5 & 5      & 5      & 5      \\
    \midrule
    \multicolumn{5}{l}{Char decoder LSTM}    \\
    -- hidden size           & 256   & 256    & 256    & 256    \\
    -- layers                & 1     & 1      & 1      & 1      \\
    -- embedding  & 128   & 128    & 128    & 128    \\
    \midrule
    \multicolumn{5}{l}{Word context encoder LSTM}    \\
    -- hidden size           & --    & --     & 768    & 768    \\
    -- layers                & --    & --     & 1      & 1      \\
    %\midrule
    % Max masked word length & --    & --     & 20     & 20     \\
    \bottomrule
  \end{tabular}
  \caption{Configurations for subnetworks unique to subword segmental LMs.}
  \label{tab:segmental_hyperparameters}
\end{table}

\section{Evaluation Tasks}
\label{appendix:evaluation}

The official 2026 BabyLM Challenge evaluation pipeline contains three types of tasks for the \textsc{Strict} and \textsc{Strict-small} tracks.

\begin{enumerate}
    \item \textbf{Zero-shot:} BLiMP \citep{warstadt-etal-2020-blimp}, BLiMP Supplement, EWoK \citep{ivanova-etal-2025-elements}, COMPS \citep{misra-etal-2023-comps}, Entity Tracking \citep{kim-schuster-2023-entity}, and the English subset of Global PIQA \citep{mrl-workshop-2025-global-piqa} (a hidden task announced shortly before the deadline).
    \item \textbf{Finetuning:} BoolQ \citep{clark-etal-2019-boolq}, MultiRC \citep{khashabi-etal-2018-looking}, RTE \citep{giampiccolo-etal-2007-third}, WSC \citep{levesque2011winograd}, MRPC \citep{dolan-brockett-2005-automatically}, QQP, and MNLI \citep{williams-etal-2018-broad}. 
    \item \textbf{Human likeness:} Reading correlations \citep{devarda2024cloze} are computed based on self-paced reading times and eye tracking data. Age-of-acquisition scores \citep{chang-bergen-2022-word} are computed by tracking word surprisal across pretraining checkpoints and comparing learning curves to child vocabulary acquisition data.    
\end{enumerate}

% For the \textsc{Multilingual} track, these zero-shot tasks additionally include ZhoBLiMP \citep{liu-etal-2026-systematic}, BLiMP-NL \citep{suijkerbuijk-etal-2025-blimp}, MultiBLiMP \citep{jumelet-etal-2026-multiblimp}, HellaSwag \citep{zellers-etal-2019-hellaswag}, Winogrande \citep{winogrande}, XCOMPS \citep{he-etal-2025-xcomps}, and XStoryCloze \citep{lin-etal-2022-shot}.

% % For the \textsc{Multilingual} track, finetuning tasks additionally include ARC \citep{arc}, Belebele \citep{bandarkar-etal-2024-belebele}, BMLama \citep{qi-etal-2023-cross}, SIB-200 \citep{adelani-etal-2024-sib}, TruthfulQA \citep{lin-etal-2022-truthfulqa},  INCLUDE \citep{romanou2025include}, and XNLI \citep{conneau-etal-2018-xnli}.

% \paragraph{3. Human-likeness tasks} The \textsc{Strict} and \textsc{Strict-small} tracks include two tests of human-likeness. 

\section{Evaluation Wrappers}
\label{appendix:evaluating_our_models}
Because our models do not output per-position logits over a fixed subword vocabulary, they cannot be evaluated directly with the official BabyLM zero-shot and human likeness evaluation pipelines. Here we describe the wrappers we implement to transform the outputs of SubSegGPT and SubSegDeBERTa (marginal probabilities) into the quantities required by each evaluation task.

\subsection{SubSegGPT} \label{appendix:evaluating_subseggpt} Zero-shot tasks compare log-probabilities of minimal pair sentences. The evaluation pipeline computes this by summing per-position log-probabilities over the tokens in a sentence. SubSegGPT computes the log-probability of a full sentence, which we transform to per-character log-probability estimates by dividing the sentence log-probability by the number of characters in a sentence. This ensures that the per-position log-probabilities summed by the evaluation pipeline add up to the true sentence log-probability, enabling valid minimal pair comparisons. 
To force the evaluation pipeline to compare full sentence log-probabilities, rather than only the log-probabilities of the differing spans between minimal pairs, we disable sentence masking in the official pipeline.\footnote{To evaluate SubSegGPT on zero-shot tasks, disable phrase masking by changing the following line to \texttt{phrase\_mask = [0 for \_ in range(len(tokens))]}:
\url{https://github.com/babylm-org/babylm-eval/blob/68cdd160f34826307e650c484904c274692e82ce/strict/evaluation_pipeline/sentence_zero_shot/dataset.py\#L89}.}

Human-likeness tasks require word-level surprisal, $-\log p(w_i \mid w_{<i})$
 which vanilla LMs compute by summing subword token log-probabilities. We instead derive this as
\begin{align}
\log p(w_i \mid w_{<i}) = \log p(w_{\leq i}) - \log p(w_{<i}),
\end{align}
where both terms on the right are computed as full sequence marginals (Equation~\ref{marginalised_formula}) using our dynamic programming algorithm.

\subsection{SubSegDeBERTa}

For MLMs, the BabyLM pipeline scores sentences with pseudo-log-likelihood \citep{salazar-etal-2020-masked}, masking each subword token in turn and summing the log-probabilities over tokens in a sentence. SubSegDeBERTa lends itself naturally to this type of evaluation, as it computes the probability of a masked word (Equation~\ref{marginalised_word_formula}), which can be used for cloze-style scoring.
For zero-shot tasks we mask each word in turn and sum the marginal of each word to compute the pseudo-log-likelihood of a sentence.  For human-likeness tasks, we mirror the BabyLM pipeline, which estimates word surprisal $-\log p(w_i \mid w_{<i})$ in MLMs by masking the target word at the end of its context, so the model conditions only on preceding words. 
For reading time evaluation, we match the official evaluation pipeline by applying multi-mask ending \citep{3737916.3738000}: appending three trailing \texttt{[MASK]} tokens to obtain a less restricted sentence continuation prediction.

\begin{table*}[h] \small
    \centering
    \renewcommand{\arraystretch}{1.3}
    \setlength{\tabcolsep}{4pt}
    \begin{tabular}{l|l|l}
    \toprule
        & \textbf{SubSegGPT} & \textbf{SubSegDeBERTa} \\
    \midrule
    \textbf{1M} & \morphemegreen{there} \morphemeplain{aren}\morphemeplain{\textquotesingle{}}\morphemeplain{t} \morphemegreen{any} \morphemeplain{gra}\morphemeplain{ham} \morphemegreen{crack}\morphemeplain{ers} \morphemegreen{sweet}\morphemegreen{ie}\morphemeplain{.} & \morphemegreen{there} \morphemeplain{aren}\morphemeplain{\textquotesingle{}}\morphemeplain{t} \morphemegreen{any} \morphemeplain{gra}\morphemeplain{ham} \morphemegreen{crack}\morphemeplain{ers} \morphemegreen{sweet}\morphemegreen{ie}\morphemeplain{.} \\
    \textbf{10M} & \morphemegreen{there} \morphemeplain{aren}\morphemeplain{\textquotesingle{}}\morphemeplain{t} \morphemegreen{any} \morphemeplain{g}\morphemeplain{raham} \morphemegreen{crack}\morphemeplain{ers} \morphemegreen{sweet}\morphemegreen{ie}\morphemeplain{.} & \morphemegreen{there} \morphemeplain{aren}\morphemeplain{\textquotesingle{}}\morphemeplain{t} \morphemegreen{any} \morphemeplain{gra}\morphemeplain{ham} \morphemegreen{crack}\morphemeplain{ers} \morphemegreen{sweet}\morphemegreen{ie}\morphemeplain{.} \\
    \textbf{50M} & \morphemegreen{there} \morphemeplain{aren}\morphemeplain{\textquotesingle{}}\morphemeplain{t} \morphemegreen{any} \morphemeplain{g}\morphemeplain{raham} \morphemegreen{crack}\morphemeplain{ers} \morphemegreen{sweet}\morphemegreen{ie}\morphemeplain{.} & \morphemegreen{there} \morphemeplain{aren}\morphemeplain{\textquotesingle{}}\morphemeplain{t} \morphemegreen{any} \morphemeplain{gra}\morphemeplain{ham} \morphemegreen{crack}\morphemegreen{er}\morphemegreen{s} \morphemegreen{sweet}\morphemegreen{ie}\morphemeplain{.} \\
    \textbf{100M} & \morphemegreen{there} \morphemeplain{aren}\morphemeplain{\textquotesingle{}}\morphemeplain{t} \morphemegreen{any} \morphemeplain{g}\morphemeplain{raham} \morphemegreen{crack}\morphemeplain{ers} \morphemegreen{sweet}\morphemegreen{ie}\morphemeplain{.} & \morphemegreen{there} \morphemeplain{aren}\morphemeplain{\textquotesingle{}}\morphemeplain{t} \morphemegreen{any} \morphemeplain{gra}\morphemeplain{ham} \morphemegreen{crack}\morphemegreen{er}\morphemegreen{s} \morphemeplain{s}\morphemeplain{weet}\morphemegreen{ie}\morphemeplain{.} \\
    \midrule
    \textbf{1M} & \morphemegreen{they}\morphemeplain{\textquotesingle{}}\morphemegreen{re} \morphemeplain{runni}\morphemeplain{ng} \morphemegreen{down} \morphemegreen{the} \morphemegreen{stair}\morphemegreen{s}\morphemeplain{.} & \morphemegreen{they}\morphemeplain{\textquotesingle{}}\morphemegreen{re} \morphemeplain{runni}\morphemeplain{ng} \morphemegreen{down} \morphemegreen{the} \morphemegreen{stair}\morphemegreen{s}\morphemeplain{.} \\
    \textbf{10M} & \morphemegreen{they}\morphemeplain{\textquotesingle{}}\morphemegreen{re} \morphemeplain{runni}\morphemeplain{ng} \morphemegreen{down} \morphemegreen{the} \morphemegreen{stair}\morphemegreen{s}\morphemeplain{.} & \morphemegreen{they}\morphemeplain{\textquotesingle{}}\morphemegreen{re} \morphemegreen{runn}\morphemegreen{ing} \morphemegreen{down} \morphemegreen{the} \morphemegreen{stair}\morphemegreen{s}\morphemeplain{.} \\
    \textbf{50M} & \morphemegreen{they}\morphemeplain{\textquotesingle{}}\morphemegreen{re} \morphemeplain{runni}\morphemeplain{ng} \morphemegreen{down} \morphemegreen{the} \morphemegreen{stair}\morphemegreen{s}\morphemeplain{.} & \morphemegreen{they}\morphemeplain{\textquotesingle{}}\morphemegreen{re} \morphemegreen{runn}\morphemegreen{ing} \morphemegreen{down} \morphemegreen{the} \morphemegreen{s}\morphemeplain{tairs}\morphemeplain{.} \\
    \textbf{100M} & \morphemegreen{they}\morphemeplain{\textquotesingle{}}\morphemegreen{re} \morphemeplain{runni}\morphemeplain{ng} \morphemegreen{down} \morphemegreen{the} \morphemegreen{stair}\morphemegreen{s}\morphemeplain{.} & \morphemegreen{they}\morphemeplain{\textquotesingle{}}\morphemegreen{re} \morphemegreen{runn}\morphemegreen{ing} \morphemegreen{down} \morphemegreen{the} \morphemegreen{s}\morphemeplain{tairs}\morphemeplain{.} \\
    \midrule
    \textbf{1M} & \morphemegreen{I}\morphemeplain{\textquotesingle{}}\morphemegreen{m} \morphemeplain{putti}\morphemeplain{ng} \morphemeplain{icing} \morphemegreen{on} \morphemegreen{them}\morphemeplain{.} & \morphemegreen{I}\morphemeplain{\textquotesingle{}}\morphemegreen{m} \morphemeplain{putti}\morphemeplain{ng} \morphemegreen{ic}\morphemegreen{ing} \morphemegreen{on} \morphemegreen{them}\morphemeplain{.} \\
    \textbf{10M} & \morphemegreen{I}\morphemeplain{\textquotesingle{}}\morphemegreen{m} \morphemeplain{putti}\morphemeplain{ng} \morphemeplain{icing} \morphemegreen{on} \morphemegreen{them}\morphemeplain{.} & \morphemegreen{I}\morphemeplain{\textquotesingle{}}\morphemegreen{m} \morphemegreen{put}\morphemeplain{ting} \morphemegreen{ic}\morphemegreen{ing} \morphemegreen{on} \morphemegreen{them}\morphemeplain{.} \\
    \textbf{50M} & \morphemegreen{I}\morphemeplain{\textquotesingle{}}\morphemegreen{m} \morphemeplain{putti}\morphemeplain{ng} \morphemeplain{icing} \morphemegreen{on} \morphemegreen{them}\morphemeplain{.} & \morphemegreen{I}\morphemeplain{\textquotesingle{}}\morphemegreen{m} \morphemegreen{put}\morphemeplain{ting} \morphemeplain{i}\morphemeplain{cing} \morphemegreen{on} \morphemegreen{them}\morphemeplain{.} \\
    \textbf{100M} & \morphemegreen{I}\morphemeplain{\textquotesingle{}}\morphemegreen{m} \morphemeplain{putti}\morphemeplain{ng} \morphemeplain{icing} \morphemegreen{on} \morphemegreen{them}\morphemeplain{.} & \morphemegreen{I}\morphemeplain{\textquotesingle{}}\morphemegreen{m} \morphemegreen{put}\morphemeplain{ting} \morphemeplain{i}\morphemeplain{cing} \morphemegreen{on} \morphemegreen{them}\morphemeplain{.} \\
    \midrule
    \textbf{1M} & \morphemegreen{did} \morphemegreen{you} \morphemegreen{find} \morphemegreen{the} \morphemeplain{littl}\morphemeplain{e} \morphemegreen{red} \morphemeplain{bicyc}\morphemeplain{le}\morphemeplain{?} & \morphemegreen{did} \morphemegreen{you} \morphemegreen{find} \morphemegreen{the} \morphemeplain{l}\morphemeplain{ittle} \morphemegreen{red} \morphemegreen{bi}\morphemegreen{cycle}\morphemeplain{?} \\
    \textbf{10M} & \morphemegreen{did} \morphemegreen{you} \morphemegreen{find} \morphemegreen{the} \morphemeplain{littl}\morphemeplain{e} \morphemegreen{red} \morphemeplain{b}\morphemeplain{icycl}\morphemeplain{e}\morphemeplain{?} & \morphemegreen{did} \morphemegreen{you} \morphemegreen{find} \morphemegreen{the} \morphemeplain{littl}\morphemeplain{e} \morphemegreen{red} \morphemegreen{bi}\morphemegreen{cycle}\morphemeplain{?} \\
    \textbf{50M} & \morphemegreen{did} \morphemegreen{you} \morphemegreen{find} \morphemegreen{the} \morphemeplain{littl}\morphemeplain{e} \morphemegreen{red} \morphemeplain{b}\morphemeplain{icycl}\morphemeplain{e}\morphemeplain{?} & \morphemegreen{did} \morphemegreen{you} \morphemegreen{find} \morphemegreen{the} \morphemeplain{littl}\morphemeplain{e} \morphemegreen{red} \morphemegreen{bi}\morphemegreen{cycle}\morphemeplain{?} \\
    \textbf{100M} & \morphemegreen{did} \morphemegreen{you} \morphemegreen{find} \morphemegreen{the} \morphemeplain{littl}\morphemeplain{e} \morphemegreen{red} \morphemeplain{b}\morphemeplain{icycl}\morphemeplain{e}\morphemeplain{?} & \morphemegreen{did} \morphemegreen{you} \morphemegreen{find} \morphemegreen{the} \morphemeplain{littl}\morphemeplain{e} \morphemegreen{red} \morphemegreen{bi}\morphemegreen{cycle}\morphemeplain{?} \\
    \midrule
    \textbf{1M} & \morphemegreen{I}\morphemeplain{\textquotesingle{}}\morphemegreen{m} \morphemeplain{dec}\morphemeplain{orati}\morphemeplain{ng} \morphemegreen{the} \morphemeplain{cooki}\morphemeplain{es}\morphemeplain{.} & \morphemegreen{I}\morphemeplain{\textquotesingle{}}\morphemegreen{m} \morphemeplain{d}\morphemeplain{ecor}\morphemeplain{ating} \morphemegreen{the} \morphemeplain{cooki}\morphemeplain{es}\morphemeplain{.} \\
    \textbf{10M} & \morphemegreen{I}\morphemeplain{\textquotesingle{}}\morphemegreen{m} \morphemeplain{de}\morphemeplain{cora}\morphemeplain{ting} \morphemegreen{the} \morphemeplain{c}\morphemeplain{ookie}\morphemegreen{s}\morphemeplain{.} & \morphemegreen{I}\morphemeplain{\textquotesingle{}}\morphemegreen{m} \morphemeplain{dec}\morphemeplain{orat}\morphemegreen{ing} \morphemegreen{the} \morphemeplain{cooki}\morphemeplain{es}\morphemeplain{.} \\
    \textbf{50M} & \morphemegreen{I}\morphemeplain{\textquotesingle{}}\morphemegreen{m} \morphemeplain{de}\morphemeplain{cora}\morphemeplain{ting} \morphemegreen{the} \morphemeplain{c}\morphemeplain{ookie}\morphemegreen{s}\morphemeplain{.} & \morphemegreen{I}\morphemeplain{\textquotesingle{}}\morphemegreen{m} \morphemeplain{d}\morphemeplain{ecor}\morphemeplain{ating} \morphemegreen{the} \morphemeplain{c}\morphemeplain{ookie}\morphemegreen{s}\morphemeplain{.} \\
    \textbf{100M} & \morphemegreen{I}\morphemeplain{\textquotesingle{}}\morphemegreen{m} \morphemeplain{dec}\morphemeplain{orati}\morphemeplain{ng} \morphemegreen{the} \morphemeplain{c}\morphemeplain{ookie}\morphemegreen{s}\morphemeplain{.} & \morphemegreen{I}\morphemeplain{\textquotesingle{}}\morphemegreen{m} \morphemeplain{d}\morphemeplain{ecor}\morphemeplain{ating} \morphemegreen{the} \morphemeplain{c}\morphemeplain{ookie}\morphemegreen{s}\morphemeplain{.} \\
    \bottomrule
    \end{tabular}
    \caption{Learned subword tokenisations of CHILDES utterances across \textsc{Strict-small} pretraining checkpoints, showing how subword boundaries evolve and highlighting subwords that correspond to morphemes.}
    \label{table_childes_examples_2}
\end{table*}

% =====================================================================
%  Reproducibility tables for the BabyLM 2026 submissions + baselines.
%  Requires \usepackage{booktabs}.  (\dagger etc. defined inline.)
%
%  Sources: model config.json on the HF Hub + the training scripts
%  (train_subseg{gpt,deberta}.sh, train_deberta_baseline.sh, and the
%  official strict-gpt2 config.yaml).  Parameter counts are the exact
%  safetensors totals of the released main-branch checkpoints.
% =====================================================================

% ------------------------------------------------------------------
% TABLE 1: Shared backbone architecture + optimisation
% ------------------------------------------------------------------
%                 Strict          Strict-small
% GPT-2           
% SubSegGPT       85h             16h

% DeBERTa         17h             2h
% SubSegDeBERTa   80h             9h

\begin{table}[t]
\centering
\small
\begin{tabular}{llcc}
\toprule
\textbf{Track} &\textbf{Model} & \textbf{Read} & \textbf{AoA} \\
\midrule
\multirow{4}{*}{\textsc{Strict}} &GPT-2&\textbf{6.93}&	-11.58		\\
&SubSegGPT 	& 1.56 & 0.00					\\
&DeBERTa	& 4.76 & 0.00							\\
&SubSegDeBERTa	&2.47&	-20.47		\\
\midrule
\multirow{4}{*}{\textsc{Strict-small}} &GPT-2&\textbf{5.63}&	-12.15		\\
&SubSegGPT	& 2.21 & 0.00					\\
&DeBERTa	& 4.17 & 0.00							\\
&SubSegDeBERTa	& 2.22 & 0.00			\\
\bottomrule
\end{tabular}
\caption{Results for human likeness tasks. \textbf{Read} reports by how much (\%) model word surprisal improves human reading time prediction. \textbf{AoA} reports correlation between model and child word acquisition. }
\label{tab:results_human_likeness}
\vspace{-0.25cm}
\end{table}

\section{Human-Likeness Results}
\label{appendix:human_likeness_results}

Neither of our models exhibits strong correlations with human psycholinguistic data, as shown in Table~\ref{tab:results_human_likeness}. 
\textbf{Read} scores are positive, so incorporating word-level surprisal from our models does slightly improve reading time prediction regression, but baseline surprisals lead to greater improvements than SubSegGPT and SubSegDeBERTa.
% reports how much adding word-level surprisal improves a  baseline (as a percentage of the maximum possible increase in $R^2$), averaged across self-paced reading time data and eye tracking data. All the models we tested improve the regression fit, but SubSegGPT and SubSegDeBERTa lead to smaller improvements than our baselines. 
% Incorporating word-level surprisal from our GPT-2 baseline achieves the best improvement among both \textsc{Strict} and x\textsc{Strict-small} models. 
\textbf{AoA} scores are zero or negative for all the models we tested, which shows that model word acquisition patterns exhibit no correlation with child acquisition data.

\end{document}